\documentclass[conference]{IEEEtran}
\IEEEoverridecommandlockouts

\usepackage{cite}
\usepackage{amsmath,amssymb,amsfonts}
\usepackage{algorithmic}
\usepackage{graphicx}
\usepackage{textcomp}
\usepackage{xcolor}
\usepackage{subcaption}
\usepackage{makecell}
\usepackage{amsmath}
\usepackage{booktabs}
\usepackage{url}
\usepackage{hyperref}
\def\BibTeX{{\rm B\kern-.05em{\sc i\kern-.025em b}\kern-.08em
    T\kern-.1667em\lower.7ex\hbox{E}\kern-.125emX}}
\begin{document}

\title{Efficient Transformer for High Resolution Image Motion Deblurring}

\author{\IEEEauthorblockN{Amanturdieva Akmaral}

\and
\IEEEauthorblockN{Muhammad Hamza Zafar}
}

\maketitle

\begin{abstract}
This paper presents a comprehensive study and improvement of the Restormer architecture for high-resolution image motion deblurring. We introduce architectural modifications that reduce model complexity by 18.4\% while maintaining or improving performance through optimized attention mechanisms. Our enhanced training pipeline incorporates additional transformations including color jitter, Gaussian blur, and perspective transforms to improve model robustness as well as new Frequency loss term. Extensive experiments on the RealBlur-R, RealBlur-J~\cite{vedaldi_real-world_2020}, and Ultra-High-Definition Motion blurred (UHDM)~\cite{zhang_mc-blur_2023} datasets demonstrate the effectiveness of our approach. The improved architecture shows better convergence behavior and reduced training time while maintaining competitive performance across challenging scenarios. We also provide detailed ablation studies analyzing the impact of our modifications on model behavior and performance. Our results suggest that thoughtful architectural simplification combined with enhanced training strategies can yield more efficient yet equally capable models for motion deblurring tasks. Code and Data Available at \url{https://github.com/hamzafer/image-deblurring}.
\end{abstract}

\begin{IEEEkeywords}
Deblurring, Transformer, Motion Blur, Deep Learning, Image Restoration
\end{IEEEkeywords}

\section{Introduction}
Image deblurring is a fundamental problem in computer vision and image processing, crucial for applications such as photography, surveillance, medical imaging, and autonomous systems~\cite{zhang_mc-blur_2023}.  The goal of image deblurring is to restore a clear, sharp image from a blurred input, where the blur may arise due to factors such as focus issues, camera shake, or rapid movement of the target~\cite{zamir_multi-stage_2021, zhang_deep_2022}. This blur severely reduces the utility of the images by obscuring crucial details, thereby degrading the performance of downstream computer vision tasks such as object detection, recognition, and segmentation. 

Traditional convolutional neural networks (CNNs) have been widely used for computer vision tasks due to their strong feature extraction capabilities~\cite{biyouki_comprehensive_2023}. However, CNNs are inherently limited by their fixed receptive fields, making it challenging for them to capture long-range dependencies effectively, particularly in high-resolution images~\cite{zamir_restormer_2022}. To overcome these limitations, recent advancements in deep learning have introduced Transformer-based architectures, which excel in capturing global context through self-attention mechanisms~\cite{vaswani_attention_2023, dosovitskiy_image_2021}. Transformers, originally developed for natural language processing, have shown remarkable success in vision tasks by modeling relationships over entire images, providing significant improvements in image quality restoration tasks~\cite{dosovitskiy_image_2021}.

In this study, we explore the use of Restormer~\cite{zamir_restormer_2022}, an efficient Transformer model specifically designed for high-resolution image restoration. Restormer introduces a multi-Dconv head transposed attention mechanism and a gated-Dconv feed-forward network, which together allow it to model long-range dependencies while maintaining computational efficiency~\cite{zamir_restormer_2022}. This Transformer model is well-suited for high-resolution images, as it captures both local and global interactions without the need for computationally prohibitive self-attention over large spatial resolutions~\cite{zamir_restormer_2022}.

Our focus is the specific type of deblurring task: motion blur. Motion blur is typically caused by the movement of the camera or objects within the scene, resulting in smeared or streaked regions across the image~\cite{lee_real-time_2024}. 

To evaluate the performance of Restormer in this contexts, we conduct experiments on a variety of benchmark datasets. For motion deblurring, we reproduced results using the RealBlur-R and RealBlur-J datasets, which provide real-world blurred images with ground truth references~\cite{vedaldi_real-world_2020}. Additionally, we curated hard positive and negative examples from these datasets to fine-tune the pre-trained model on the RealBlur~\cite{vedaldi_real-world_2020}. This fine-tuning aims to further improve the model’s ability to generalize across different blur types and intensities, especially in real world conditions. Furthermore, we retrained the baseline model with provided parameters to validate reproducibility of the results and conduct ablation studies for further improvements. To enhance the baseline model, we conducted comprehensive research to include modifications to the architecture and the training process. Finally, to evaluate our additions we benchmark the described models on a novel dataset provided in~\cite{zhang_mc-blur_2023} called Ultra-High-Definition Motion blurred set (UHDM) for motion blur tasks. To the best of our knowledge, this dataset has not yet been evaluated using the Restormer model.

In this report, we describe our methodology for fine-tuning Restormer on deblurring datasets and assess the effectiveness of our approach in enhancing the model for restoring image clarity across different blurring conditions.

\section{Related Work}

This study conducts a comprehensive comparative analysis of the primary methodology against existing approaches in the field. Through critical evaluation, both strengths and limitations of various methodologies are examined, leading to the identification of significant research opportunities and knowledge gaps. The following publications were selected for comparative analysis:

\begin{itemize}
    \item \textbf{Uformer~\cite{wang_uformer_2022}:} The paper presents a Transformer-based architecture for image restoration tasks, featuring a U-shaped hierarchical design enhanced with two key components. The first is the Locally-enhanced Window (LeWin) Transformer block, which employs non-overlapping window-based self-attention to efficiently capture long-range dependencies while reducing computational complexity. The second is a learnable multi-scale restoration modulator that adaptively adjusts features across multiple decoder layers to enhance restoration quality.  
    \item \textbf{Stripformer~\cite{Tsai_Peng_Lin_Tsai_Lin_2022}:} Authors introduce a novel transformer-based architecture for image deblurring that efficiently handles region-specific blur patterns through strip-based attention mechanisms. The advancement includes two key components: intra-strip attention for pixel-wise feature dependencies within horizontal and vertical strips, and inter-strip attention for capturing global region-wise correlations. 
    \item \textbf{Multi-scale Cubic-Mixer~\cite{zheng_uhd_2022}:} This study proposes Multi-scale Cubic-Mixer, a deep network architecture for image deblurring that operates without self-attention mechanisms to achieve computational efficiency. The model's novelty lies in its frequency-domain approach, where it processes blurred images through Fast Fourier Transform to work with both real and imaginary components of the Fourier coefficients. By operating in the frequency domain and utilizing a three-dimensional (Channel, Width, Height) MLP structure, the network is able to captures both long-range dependencies and local features from blurred images. 
    \item \textbf{Adversarial Promoting Learning (APL)~\cite{Zhao_Wei_He_Lu_2022}: } A novel framework that jointly handles defocus detection and deblurring without requiring pixel-level annotations or paired deblurring ground truth is presented. The approach is based on the complementary nature of these tasks - defocus detection guides deblurring by segmenting focused areas, while effective deblurring necessitates accurate defocus detection. The framework consists of three key components: a defocus detection generator ($G_{ws}$) that produces detection maps to segment focused and defocused regions, a self-referenced deblurring generator ($G_{sr}$) that utilizes the focused areas as references to restore defocused regions, and a discriminator that enables adversarial optimization. Both generators are trained alternately through adversarial learning against unpaired fully-clear images, where $G_{sr}$ aims to produce convincing deblurred results while $G_{ws}$ is driven to generate accurate detection maps to guide the deblurring process. 
    \item \textbf{Test-time Local Converter (TLC)~\cite{chu_improving_2022}: } Authors propose a TLC approach to address the train-test inconsistency in image restoration networks that use global operations(like global average pooling and normalization). The key insight is that these operations behave differently during training with patches versus inference with full images, leading to shifts in feature distribution that degrade performance. TLC converts these global operations to local ones during inference by aggregating features within local spatial windows rather than across the entire image. 
\end{itemize}

\subsection{Positive Aspects}
\begin{itemize}
    \item Restormer is a versatile model applicable to various image restoration tasks (e.g., denoising, and deblurring). Its design allows it to handle different restoration tasks within the same framework, while many other methods are more specialized.
    \item Restormer's architecture is specifically tailored for high-resolution images, where traditional CNNs or other Transformers would struggle with the larger spatial dimensions. 
\end{itemize}

\subsection{Negative Aspects}
\begin{itemize}
    \item Restormer is optimized as a general-purpose restoration tool and lacks explicit mechanisms to address defocus-specific artifacts. Models like APL~\cite{Zhao_Wei_He_Lu_2022}, which focus on defocus detection and targeted deblurring, could outperform it in scenarios with significant defocus blur.
    \item Unlike frameworks that integrate defocus blur detection, Restormer lacks built-in mechanisms to identify and focus on blurred regions, potentially leading to over- or under-processing of different image areas.
\end{itemize}

\subsection{Research Gap}

Restormer model's complexity leads to substantial memory and computational demands, especially during inference on devices with limited resources. Models like Stripformer, which employ simpler, strip-based attention, demonstrate that reducing model complexity without sacrificing performance is feasible. A more streamlined architecture could make Restormer more accessible for real-time applications and edge devices. It's substantial training time, is a significant limitation in practical settings. In addition, incorporating region-specific blur detection could be an another area of research. 

\section{Restormer Overview}
This section covers the details of the proposed method in ~\cite{zamir_restormer_2022}.

\subsection{Detailed Summary of Restormer Model~\cite{zamir_restormer_2022}}

The Restormer model introduces an efficient Transformer architecture specifically designed for high-resolution image restoration tasks by addressing computational bottlenecks. This is achieved through innovative modifications to the multi-head self-attention (SA) layer and the adoption of a multiscale hierarchical module, which reduces computational demands compared to traditional single-scale networks.

\begin{figure*}[ht]
    \centering
    \includegraphics[width=0.8\linewidth]{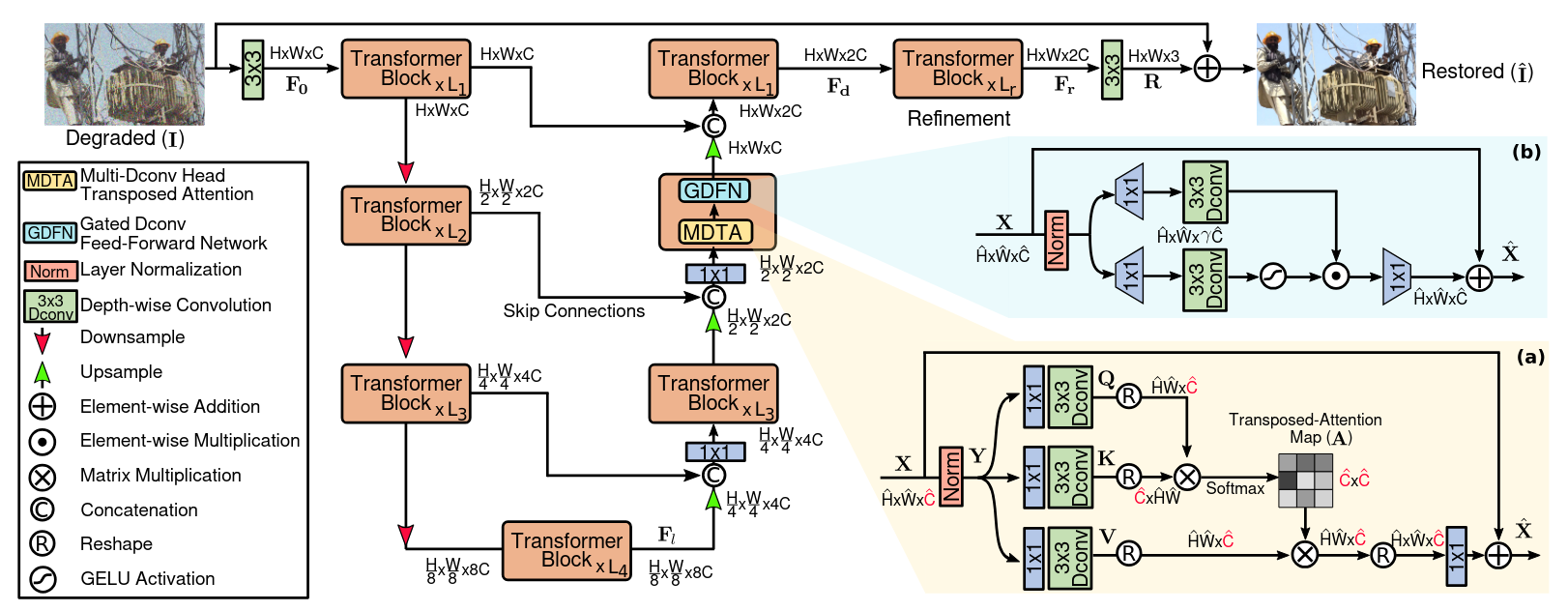}
    \caption{Architecture of Restormer for high-resolution image restoration from \cite{zamir_restormer_2022} (a) Multi-Dconv Head Transposed Attention (MDTA) and (b) Gated-Dconv Feed-Forward Network (GDFN)}
    \label{fig:restormer}
\end{figure*}

\subsection{Pipeline Overview~\cite{zamir_restormer_2022}}

Restormer follows a 4-level symmetric encoder-decoder architecture (see Figure~\ref{fig:restormer}). Initially, a degraded input image $I \in \mathbb{R}^{H \times W \times 3}$ is processed using a convolution layer to extract low-level feature embeddings $F_0 \in \mathbb{R}^{H \times W \times C}$. These features are passed through the encoder-decoder pipeline, which progressively reduces spatial dimensions while increasing the channel capacity. Each level of the encoder-decoder architecture is composed of multiple Transformer blocks, with the number of blocks increasing progressively from the upper levels to the lower levels, ensuring computational efficiency.
\begin{itemize}
    \item \textbf{Encoder:} Downsampling is performed hierarchically to extract latent features, while the number of Transformer blocks increases progressively across levels to enhance efficiency.
    \item \textbf{Decoder:} The decoder reconstructs high-resolution representations from the low-resolution latent features $F_l \in \mathbb{R}^{H/8 \times W/8 \times 8C}$ using upsampling techniques.
\end{itemize}
To manage downsampling and upsampling, the architecture employs pixel-unshuffle and pixel-shuffle operations~\cite{shi_real-time_2016}, respectively. Encoder features are concatenated with decoder features through skip connections~\cite{ronneberger_u-net_2015} to enhance the recovery of fine details. Post-concatenation, a $1 \times 1$ convolution is applied at all levels (except the top) to halve the channel count.

The refined feature maps $F_d$ are enriched at a high spatial resolution in the final refinement stage, improving structural and textural detail preservation. The output residual image $R \in \mathbb{R}^{H \times W \times 3}$ is combined with the input to produce the restored image: $\hat{I} = I + R$.

The Restormer architecture introduces two primary innovations in its Transformer block: 

\begin{enumerate}
    \item \textbf{Multi-Dconv Head Transposed Attention (MDTA):}  
    MDTA redefines self-attention (SA) by computing cross-covariance across channels instead of spatial dimensions, resulting in a global attention map with linear complexity. It involves depth-wise convolutions to enhance local context before generating the attention map. More precisely, from a layer-normalized tensor $Y \in \mathbb{R}^{\hat{H} \times \hat{W} \times \hat{C}}$, the MDTA module generates query ($Q$), key ($K$), and value ($V$) projections, enriched with local context. This is achieved using $1 \times 1$ point-wise convolutions ($W_p^{(\cdot)}$) for cross-channel aggregation and $3 \times 3$ depth-wise convolutions ($W_d^{(\cdot)}$) for encoding spatial context: 
    \[Q = W_d^Q W_p^Q Y, \quad K = W_d^K W_p^K Y, \quad V = W_d^V W_p^V Y.\]
    The query and key are reshaped such that their dot-product interaction produces a transposed-attention map $A \in \mathbb{R}^{\hat{C} \times \hat{C}}$, avoiding the large regular attention map of size $\mathbb{R}^{\hat{H}\hat{W} \times \hat{H}\hat{W}}$. The MDTA process is defined as:
    \[\hat{X} = W_p \cdot \text{Attention}(\hat{Q}, \hat{K}, \hat{V}) + X,\]
    \[\text{Attention}(\hat{Q}, \hat{K}, \hat{V}) = \hat{V} \cdot \text{Softmax} \left( \frac{\hat{K} \cdot \hat{Q}}{\alpha} \right),\]
    where $X$ and $\hat{X}$ are the input and output feature maps. Projections $\hat{Q} \in \mathbb{R}^{\hat{H}\hat{W} \times \hat{C}}$, $\hat{K} \in \mathbb{R}^{\hat{C} \times \hat{H}\hat{W}}$, and $\hat{V} \in \mathbb{R}^{\hat{H}\hat{W} \times \hat{C}}$ are obtained by reshaping tensors from the original $\mathbb{R}^{\hat{H} \times \hat{W} \times \hat{C}}$. 
    Here, $\alpha$ is a learnable scaling parameter for controlling the dot-product magnitude, and channels are divided into multiple heads to learn attention maps in parallel, similar to conventional multi-head self-attention (SA)~\cite{vaswani_attention_2023}.
    
    \item \textbf{Gated-Dconv Feed-Forward Network (GDFN):}
    The GDFN module uses the element-wise product of two parallel transformations — one linear and the other activated with the GELU~\cite{hendrycks_gaussian_2023} non-linearity — to control the flow of information through the network. Depth-wise convolutions are applied to encode spatially neighboring pixel information, further enriching the local context.
    
    Given an input tensor $X \in \mathbb{R}^{\hat{H} \times \hat{W} \times \hat{C}}$, the GDFN is defined as:
    \[  \hat{X} = W_p^0 \cdot \text{Gating}(X) + X,\]
    \[\text{Gating}(X) = \phi\left(W_d^1 W_p^1 (\text{LN}(X))\right) \odot W_d^2 W_p^2 (\text{LN}(X)),\]
    where $\odot$ represents element-wise multiplication, $\phi$ is the GELU activation function, and $\text{LN}$ denotes layer normalization.
    The GDFN regulates the flow of information between hierarchical levels, enabling each level to emphasize fine details. Unlike MDTA, which enriches features with contextual information, GDFN focuses on controlled feature transformation.
\end{enumerate}

\subsection{Implementation Details~\cite{zamir_restormer_2022}}

The model is trained progressively, starting with smaller image patches and bigger batch size in earlier epochs and gradually transitioning to larger patches and smaller batch size $-$ a process called progressive training. The Restormer model is trained with the objective of restoring sharpness and fine details, minimizing perceptual and structural differences between deblurred images and ground truth. 

Key \textbf{hyperparameters and configurations} used in training Restormer~\cite{zamir_restormer_2022} include the following:

\begin{itemize}
    \item \textbf{Transformer Architecture}: 
    The model employs a hierarchical structure with the following configurations across levels:
    \begin{itemize}
        \item \textbf{Number of Transformer Blocks}: [4, 6, 6, 8] for levels 1 to 4.
        \item \textbf{Attention Heads in MDTA}: [1, 2, 4, 8], progressively increasing to capture richer contextual information.
        \item \textbf{Number of Channels}: [48, 96, 192, 384], allowing for greater feature representation at deeper levels.
    \end{itemize}
    The refinement stage contains 4 additional Transformer blocks, and the channel expansion factor in GDFN is set to $\gamma = 2.66$.

    \item \textbf{Learning Rate}: 
    The initial learning rate is set to $3 \times 10^{-4}$ and is gradually reduced to $10^{-6}$ using a cosine annealing schedule~\cite{loshchilov_sgdr_2017}. This helps prevent overfitting and stabilizes convergence during training.

    \item \textbf{Optimizer}: 
    The AdamW optimizer~\cite{loshchilov_decoupled_2019} is used with parameters $\beta_1 = 0.9$, $\beta_2 = 0.999$, and a weight decay of $1 \times 10^{-4}$. 

    \item \textbf{Loss Function}: 
    The $L_1$ loss is employed for training, ensuring pixel-wise precision and minimizing absolute differences between predictions and ground truth. It is  also known as mean absolute error (MAE), which is defined as: \[
\mathcal{L}_{1} = \frac{1}{N} \sum_{i=1}^N \left| \hat{y}_i - y_i \right|,
\]
where \(N\) is the number of pixels, \(\hat{y}_i\) represents the predicted pixel value, and \(y_i\) is the corresponding ground truth value. This loss penalizes deviations linearly, encouraging the model to minimize the absolute differences between predictions and targets.

    \item \textbf{Training Schedule}: 
    Models are trained for 300K iterations, starting with a patch size of $128 \times 128$ and a batch size of 64. Progressive learning is applied by updating patch size and batch size pairs as follows: 
\[ \{(160^2, 40), (192^2, 32),  (256^2, 16), (320^2, 8), (384^2, 8)\},\]
    at iterations \{92K, 156K, 204K, 240K, 276K\}, respectively.

    \item \textbf{Data Augmentation}: 
    Horizontal and vertical flips are applied to enhance the diversity of the training dataset.
\end{itemize}

\subsection{Evaluation Metrics~\cite{zamir_restormer_2022}}

Authors in~\cite{zamir_restormer_2022} compute evaluation metrics using standard practices to ensure comparability with existing methods. Below are the details of the metrics:

\paragraph{PSNR/SSIM in Y Channel:} 
PSNR and SSIM scores are computed using the Y channel of the YCbCr color space~\cite{zamir_restormer_2022}. This focuses the evaluation on luminance information, which is critical for perceptual quality, while ignoring chrominance channels (Cb, Cr) that may introduce biases.

The PSNR measures the ratio between the maximum possible intensity of an image and the mean squared error (MSE) between the predicted image $\hat{I}$ and the ground truth $I$. It is defined as:
\[
\text{PSNR} = 10 \cdot \log_{10}\left(\frac{\text{MAX}^2}{\text{MSE}}\right)\text{~\cite{psnr}},
\]
where:
\begin{itemize}
    \item $\text{MAX}$ is the maximum possible pixel value (e.g., 255 for 8-bit images),
    \item $\text{MSE}$ is the mean squared error, given by:
    \[
    \text{MSE} = \frac{1}{N} \sum_{i=1}^N \left(\hat{I}_i - I_i\right)^2,
    \]
    with $N$ as the total number of pixels.
\end{itemize}

Higher PSNR values ($>$ 30dB) indicate better image restoration quality.

SSIM evaluates the similarity between two images by considering luminance, contrast, and structural information. It is computed as:
\[
\text{SSIM}(I, \hat{I}) = \frac{(2\mu_I\mu_{\hat{I}} + C_1)(2\sigma_{I\hat{I}} + C_2)}{(\mu_I^2 + \mu_{\hat{I}}^2 + C_1)(\sigma_I^2 + \sigma_{\hat{I}}^2 + C_2)}\text{~\cite{ssim}},
\]
where:
\begin{itemize}
    \item $\mu_I$ and $\mu_{\hat{I}}$ are the mean pixel intensities of $I$ and $\hat{I}$,
    \item $\sigma_I^2$ and $\sigma_{\hat{I}}^2$ are the variances of $I$ and $\hat{I}$,
    \item $\sigma_{I\hat{I}}$ is the covariance between $I$ and $\hat{I}$,
    \item $C_1 = (k_1 \cdot \text{MAX})^2$ and $C_2 = (k_2 \cdot \text{MAX})^2$ are constants to stabilize the equation, where $k_1 \ll 1$ and $k_2 \ll 1$.
\end{itemize}

SSIM ranges from 0 to 1, with values closer to 1 indicating higher structural similarity.
\paragraph{LPIPS:} 
The Learned Perceptual Image Patch Similarity (LPIPS) metric is calculated to evaluate perceptual similarity between restored and ground truth images. Authors use the implementation from the LPIPS library, with a pre-trained AlexNet backbone. The metric is computed as:

\[
\text{LPIPS}(x, y) = \sum_l \frac{1}{H_l W_l} \sum_{h,w} \| \phi_l(x)_{h,w} - \phi_l(y)_{h,w} \|_2^2 \text{~\cite{lpips}},
\]

where:
\begin{itemize}
    \item \(x\) and \(y\) are the restored and ground truth image patches,
    \item \(\phi_l\) represents the feature map from layer \(l\) of AlexNet,
    \item \(H_l\) and \(W_l\) are the dimensions of the feature map at layer \(l\).
\end{itemize}

\section{Customizations}

This section outlines the customizations and enhancements implemented to improve the Restormer model's effectiveness in motion blur. These modifications were designed to overcome challenges identified during the evaluation of the paper.

Upon evaluating the Restormer models on real-world dataset~\cite{vedaldi_real-world_2020}, we observed that the baseline configurations struggled to handle anomalies involving color differences. To address this challenge, we introduced additional transformations during training to improve model robustness and performance. 

The baseline Restormer training pipeline includes only horizontal and vertical flips, and no transformations related to color changes are applied.  To bridge this gap, we introduced color jitter, Gaussian blur, brightness and contrast adjustment transformations, enabling the model to better handle changes in illumination, sharpness, and color intensity. These transformations simulate real-world conditions, such as varying illumination, sharpness, and color intensity, allowing the model to better handle diverse visual scenarios~\cite{Yoshihara_Fukiage_Nishida_2023}. Perspective transforms were also applied to mimic geometric distortions, further increasing the variability and representativeness of the training data.

\begin{figure}[ht]
       \centering
       \includegraphics[width=\linewidth]{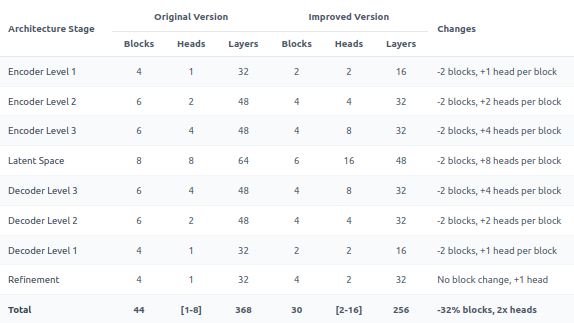}
       \caption{Detailed architecture stage comparison across encoder, latent space, decoder, and refinement blocks.}
       \label{fig:arch}
\end{figure}

\begin{figure}[ht]
       \centering
       \includegraphics[height = 4cm,width=\linewidth]{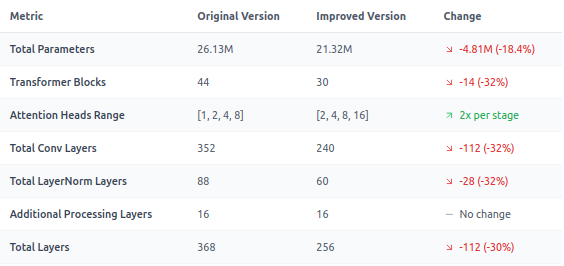}
       \caption{Model architecture changes showing reduction in parameters and layers while increasing attention heads.}
       \label{fig:changes}
\end{figure}

In parallel, we focused on optimizing the architecture to reduce model complexity while maintaining, and in some cases improving, performance. We decreased the number of layers and Transformer blocks in both the encoder and decoder, which led to a reduction in the number of parameters (by 18.4\%) and computational cost (see Figure~\ref{fig:changes}). To compensate for the reduction in layers, we doubled the number of attention heads per stage, allowing the model to better capture global and local features in a computationally efficient manner. Overall changes made to the model can observed from Figures~\ref{fig:arch} and~\ref{fig:changes}. These changes were inspired by the work done in ~\cite{vaswani_attention_2023}. Authors claim that attention mechanisms are scalable and adaptable to vision tasks, even with reduced computational overhead~\cite{vaswani_attention_2023}.

For evaluation, we incorporated color difference metric - deltaE$_{2000}$ as expressed in~\cite{Sharma_Wu_Dalal_2005} to complement standard pixel-level metrics like PSNR and SSIM. It is defined as:

\[
\Delta E_{00} \;=\; \sqrt{%
  \begin{aligned}
    &\left( \frac{\Delta L'}{k_L S_L}\right)^2 
     + \left(\frac{\Delta C'}{k_C S_C}\right)^2
     + \left(\frac{\Delta H'}{k_H S_H}\right)^2 \\
    &\quad
     +\,R_T \,\frac{\Delta C'}{k_C S_C}\,\frac{\Delta H'}{k_H S_H}
  \end{aligned}
}.
\]

\noindent
where:
\[
\begin{aligned}
\Delta L' = L_2 - L_1,  \text{(lightness difference)} \\
\Delta C' = C_2' - C_1', \text{(chroma difference)} \\
\Delta H' = 2 \sqrt{C_1' C_2'} \sin \left( \frac{\Delta h'}{2} \right), \text{(hue difference)} \\
R_T = -2 \sqrt{\frac{C'_\text{avg}{^7}}{C'_\text{avg}{^7} + 25^7}} \sin\left(60^\circ e^{-\left(\frac{h'_\text{avg} - 275}{25}\right)^2}\right), \\\text{(rotation term)}
\end{aligned}
\]

\noindent
Scaling factors \(S_L\), \(S_C\), and \(S_H\) are functions of \(L_\text{avg}\), \(C_\text{avg}\), and \(h_\text{avg}\), and are adjusted to account for perceptual uniformity differences.

These metrics provide a quantitative measure of the model's ability to restore accurate color fidelity by comparing perceptual color differences between restored images and ground truth. This addition offers a more comprehensive assessment of the model's performance, particularly in scenarios where precise color reproduction is critical.

Additionally, the loss function was modified to improve the model's ability to restore both spatial and frequency-domain details. Alongside the baseline pixel-wise  $\mathcal{L}_{1}$ loss, we introduced a frequency-domain loss inspired by Fourier transform analysis as discussed in~\cite{Benjdira_Ali_Koubaa_2023}. The combined loss is defined as:
\[
\mathcal{L}_{\text{total}} = \mathcal{L}_{\text{pixel}} + \lambda \mathcal{L}_{\text{freq}},
\]

where \( \mathcal{L}_{\text{pixel}} \) represents the pixel-level \( L_1 \) loss, and \( \mathcal{L}_{\text{freq}} \) ensures the preservation of high-frequency details critical for visual quality. The frequency loss is defined as:

\[
\mathcal{L}_{\text{freq}} = \frac{1}{N} \sum_{i=1}^{N} \lVert |\mathcal{F}(\hat{I}_i)| - |\mathcal{F}(I_i)| \rVert_1,
\]

where \( \mathcal{F} \) denotes the Fourier transform, \( |\mathcal{F}(I)| \) is the magnitude of the Fourier transform, \( \hat{I}_i \) and \( I_i \) are the restored and ground truth images respectively, and \( N \) is the total number of images in the batch. . This term emphasizes minimizing the discrepancy between the frequency components of the restored and ground truth images, enhancing the model's ability to recover sharp edges and fine textures. The weighting factor $\lambda$ was set to  0.1 in our experiments.

\section{Datasets}

\subsection{GoPro~\cite{Nah_Kim_Lee_2017}}

The GoPro dataset comprises 3,214 pairs of blurry and sharp images with a resolution of 1280×720~\cite{Nah_Kim_Lee_2017}. Typically, these are divided into 2,103 images for training and 1,111 images for testing~\cite{Nah_Kim_Lee_2017}. However, for training Restormer~\cite{zamir_restormer_2022} authors utilized the whole dataset. The dataset is created using videos captured with a GoPro camera, where multiple consecutive frames are averaged to produce various blurred images. The mid-frame of each sequence serves as the ground-truth image corresponding to the associated synthetic blurred image. Figure~\ref{fig:gopro} in Appendix~\ref{appendix} illustrates several examples of blurred images from the GoPro dataset.

\subsection{RealBlur~\cite{vedaldi_real-world_2020}}

The RealBlur dataset consists of 4,738 pairs of sharp and blurred images captured using a dual-camera system with synchronized Sony A7RM3 cameras and wide-angle lenses. It includes images of resolution 680×773, processed to reduce noise and ensure alignment. The dataset is divided into 3,758 image pairs for training and 980 pairs for testing. Unlike synthetic datasets, which blend sharp frames to generate blur, RealBlur captures authentic motion blur under diverse indoor and outdoor scenes, including low-light conditions. Two variants of the dataset are provided: RealBlur-R (raw format) and RealBlur-J (JPEG format). Postprocessing involves denoising, geometric alignment, and photometric alignment to create high-quality training and evaluation data. Examples from the datasets are shown in Figures~\ref{fig:realblur} in Appendix~\ref{appendix}.

\subsection{UHDM~\cite{zhang_mc-blur_2023}}

The UHDM subset within the MC-Blur dataset comprises 8,000 training images and 2,000 testing images, all at 4K–6K resolution. The dataset is designed to address challenges in deblurring Ultra-High-Definition (UHD) images by incorporating large blur kernels with sizes ranging from 111×111 to 191×191 pixels. These kernels are generated using 3D camera trajectories and convolved with sharp UHD images, providing realistic and varied motion blur. This subset emphasizes restoring fine details required for UHD image deblurring, presenting a significant challenge for current deblurring algorithms. It is the first large-scale UHD motion-blurred dataset, catering to the increasing prevalence of high-definition cameras in real-world applications. Figure~\ref{fig:uhdm} in Appendix~\ref{appendix} presents example images from the set. 

\section{Experiments and Results}

\subsection{ Experimental Setup and Implementation }
Our experimental evaluation was conducted using PyTorch framework on NVIDIA RTX 4090 GPU. To ensure comprehensive analysis, we implemented four distinct experimental configurations. First, we reproduced the baseline results using the authors' provided model checkpoints. Second, we fine-tuned provided checkpoint on the benchmark dataset, authors used in their paper to test their models. Second, we trained the model from scratch following the original specifications to validate reproducibility. Finally, we evaluated our improved model incorporating the architectural modifications and training enhancements described in Figures~\ref{fig:arch} and~\ref{fig:changes}.

We conducted evaluations on multiple benchmark datasets to assess model performance across different scenarios. The RealBlur-R and RealBlur-J datasets from~\cite{vedaldi_real-world_2020}, each containing 980 image pairs, provided real-world blurred images with corresponding ground truth references. Additionally, we utilized the UHDM dataset~\cite{zhang_mc-blur_2023} comprising 2000 high-resolution image pairs to evaluate performance on higher resolution imagery. To analyze model behavior on challenging cases, we identified hard positives and negatives using a PSNR threshold between 20dB and 30dB provided in Table~\ref{tab:hard_positives_negatives}.

\begin{table}[ht]
\centering
\caption{Performance comparison based on hard positives and hard negatives across different datasets with the threshold of PSNR between 20dB and 3dB.}

\label{tab:hard_positives_negatives}
\resizebox{\columnwidth}{!}{%
\begin{tabular}{|l|l|c|c|c|}
\hline
\textbf{Dataset} & \textbf{Model} & \textbf{Hard Positives $\uparrow$} & \textbf{Hard Negatives $\downarrow$} & \textbf{Total Count} \\
\hline
RealBlur-R & Improved Model (ours) & 705 & \underline{17} & 980 \\
 & Fine-tuned with RealBlur-R  & \textbf{866} & \textbf{2} &  \\
 & Model checkpoint & 699 & 18 & \\
 & Reproduced Model (ours) & \underline{706} & 18 & \\ \hline
RealBlur-J & Improved Model (ours) & 164 & 27 &  980\\
 & Fine-tuned with RealBlur-J & \textbf{499} & \textbf{2} &  \\
 & Model checkpoint & \underline{166} & 27 &  \\
 & Reproduced Model (ours) & 164 & 27 & \\ \hline
UHDM & Improved Model & 4 & \textbf{598} & 2000 \\
 & Fine-tuned with RealBlur-J  & 4 & 1000 &  \\
 & Fine-tuned with RealBlur-R  & \textbf{5} & 632 &  \\
 & Model checkpoint & 4 & 608 & \\
 & Reproduced Model (ours) & 4 & \underline{604} &  \\
\hline
\end{tabular}%
}
\end{table}

\subsection{Motion Blur Baseline Model Results}

To evaluate the performance of the Restormer model on motion blur tasks, we utilized the pre-trained model checkpoints provided by the authors to reproduce their results. These checkpoints were tested on the \textbf{RealBlur-J} and \textbf{RealBlur-R} datasets~\cite{vedaldi_real-world_2020} separately, aligning with the evaluation protocol described in the original paper. Additionally, we trained the model from scratch using the \textbf{GoPro dataset}~\cite{Nah_Kim_Lee_2017}, adhering to the hyperparameters and training settings detailed in the paper, to assess the reproducibility of the reported results.

After having the reproduced model, we observed slight deviations in performance compared to the checkpoint provided by the authors. These deviations (see Tables~\ref{tab:rbr1}, \ref{tab:rbr2}, \ref{tab:rbj1} and~\ref{tab:rbj2}) can be attributed to the \textit{random initialization of the network} and the lack of detailed information on how the data was split during training, which could result in differences in the dataset used.

\begin{table}[ht]
\centering
\caption{Performance of the models trained on 8 GPUs on RealBlur RAW~\cite{vedaldi_real-world_2020}.}
\label{tab:rbr1}
\resizebox{\columnwidth}{!}{%
\begin{tabular}{l|c|c|c|c|c}
\hline
\textbf{Model} & \textbf{PSNR } $\uparrow$ & \textbf{SSIM } $\uparrow$ & \textbf{MAE } $\downarrow$ & \textbf{LPIPS } $\downarrow$ & \textbf{DeltaE } $\downarrow$ \\
\hline
{Results in the paper} &  36.190 & 0.957 & \text{NA} & \text{NA} & \text{NA} \\
{Model checkpoint} & 33.998 & 0.946 & 0.009 & 0.042 & 0.839 \\
\hline
\end{tabular}%
}
\end{table}

\begin{table}[ht]
\centering
\caption{Performance of the models trained on 1 GPU on RealBlur RAW~\cite{vedaldi_real-world_2020}.}
\label{tab:rbr2}
\resizebox{\columnwidth}{!}{%
\begin{tabular}{l|c|c|c|c|c}
\hline
\textbf{Model} & \textbf{PSNR } $\uparrow$ & \textbf{SSIM } $\uparrow$ & \textbf{MAE } $\downarrow$ & \textbf{LPIPS } $\downarrow$ & \textbf{DeltaE } $\downarrow$ \\
\hline
{Reproduced model (ours)} & 33.685 & 0.938 & 0.010 & 0.063 & 0.866 \\
{Fine-tuned on RealBlur-R} & \textbf{37.194} & \textbf{0.964} & \textbf{0.007} & \textbf{0.034} & \textbf{0.685} \\
{Improved Model (ours)} & \underline{33.997} & \underline{0.945} & \underline{0.009} & \underline{0.051} & \underline{0.856} \\
\hline
\end{tabular}%
}
\end{table}

\begin{table}[ht]
\centering
\caption{Performance of the models trained on 8 GPUs on RealBlur JPEG~\cite{vedaldi_real-world_2020}}
\label{tab:rbj1}
\resizebox{\columnwidth}{!}{%
\begin{tabular}{l|c|c|c|c|c}
\hline
\textbf{Model} & \textbf{PSNR } $\uparrow$ & \textbf{SSIM } $\uparrow$ & \textbf{MAE } $\downarrow$ & \textbf{LPIPS } $\downarrow$ & \textbf{DeltaE } $\downarrow$ \\
\hline
Results in the paper & 28.960 & 0.879 & \text{NA} & \text{NA} & \text{NA} \\
Model checkpoint & 26.628 & 0.824 & 0.029 & 0.114 & 2.607 \\
\hline
\end{tabular}%
}
\end{table}

\begin{table}[ht]
\centering
\caption{Performance of the models trained on 1 GPU on RealBlur JPEG~\cite{vedaldi_real-world_2020}}
\label{tab:rbj2}
\resizebox{\columnwidth}{!}{%
\begin{tabular}{l|c|c|c|c|c}
\hline
\textbf{Model} & \textbf{PSNR } $\uparrow$ & \textbf{SSIM } $\uparrow$ & \textbf{MAE } $\downarrow$ & \textbf{LPIPS } $\downarrow$ & \textbf{DeltaE } $\downarrow$ \\
\hline
Reproduced model (ours) & 26.570 & 0.821 & 0.029 & 0.127 & 2.603 \\
Fine-tuned on RealBlur-J & \textbf{29.945} & \textbf{0.896} & \textbf{0.021} & \textbf{0.079} & \textbf{1.679} \\
Improved Model (ours) & \underline{26.615} & \underline{0.825} & \underline{0.029} & \underline{0.121} & \underline{2.602} \\
\hline
\end{tabular}%
}
\end{table}

Moreover, the authors trained their model using \textbf{progressive training on 8 GPUs}, allowing for larger batch sizes and more efficient training. Specifically, their progressive training setup started with a patch size of $128 \times 128$ and a batch size of 64, which was progressively updated through the following configurations:
\[\{(160,40), (192,32), (256,16), (320,8), (384,8)\},\]
at iterations $\{92K, 156K, 204K, 240K, 276K\}$.
In contrast, due to \textbf{GPU memory constraints}, our setup used a smaller batch size for progressive training starting with a patch size of $128 \times 128$ and a batch size of 8, and
\[\{ (160,4), (192,4), (256,2), (320,1), (320,1)\}.\]

Despite these limitations, our training approach remained consistent with the overall methodology described by the authors, ensuring a fair comparison of results.

Furthermore, we observed that even the authors’ provided checkpoints deviate significantly from the results reported in their paper. This can be observed in Tables~\ref{tab:rbr1} and~\ref{tab:rbj1}. This discrepancy likely stems from a lack of detailed information regarding data preparation, such as how training and validation sets were split.

These differences in training configurations, data preparation, and resource availability underscore the challenges of achieving the reported state-of-the-art results, especially for high-resolution tasks like motion blur restoration.

\subsection{Fine-tuned Model Results}
Fine-tuning the baseline model on specific datasets yielded substantial improvements in performance metrics. From the Table~\ref{tab:rbr2} we can see significant 3.509 dB gain in PSNR. Notably, the DeltaE metric improved from 0.866 to 0.685, demonstrating enhanced color fidelity. RealBlur-J dataset follows the same trend, according to the Table~\ref{tab:rbj2}. 

The analysis of hard examples provides particularly interesting insights into models behavior. As shown in Figures~\ref{fig:blurRbad} and~\ref{fig:blurJbad} in the Appendix~\ref{appendix}, we identified challenging cases from the RealBlur-R and -J datasets. These examples demonstrate severe motion blur combined with low lighting conditions especially in Figure~\ref{fig:blurRbad} (Appendix~\ref{appendix}). All of the models struggle with these images, producing artifacts in regions of high frequency detail and failing to properly restore clarity. However, fine-tuned model shows great improvement even visually. The same can be observed in hard positive examples (Figures~\ref{fig:blurRgood} and~\ref{fig:blurJgood}, see Appendix~\ref{appendix}), where all models perform reasonably well, but subtle differences emerge and fine-tuned models present superior results both quantitatively and visually. Notably, the models fine-tuned on RealBlur datasets performed relatively poorly on UHDM, suggesting potential overfitting to their respective domains (see Table~\ref{tab:performance_comparison}). 

\subsection{Improved Model Results}

To evaluate the results, we benchmarked the models on a new dataset UHDM~\cite{zhang_mc-blur_2023} containing 2000 high resolution images with different levels of blur. When comparing performance across different model variants (see Table~\ref{tab:performance_comparison}), the improved model achieved the highest PSNR, showcasing marginal improvements over the baseline checkpoint and the reproduced model. The improved model's better performance can be attributed to its architectural modifications and enhanced training regime. This is particularly evident in the hard example analysis in Table~\ref{tab:hard_positives_negatives}, where the improved model reduced the number of hard negatives to 598 compared to the baseline's 604, checkpoint's 608 and fine-tuned models' higher counts (up to 1000 for RealBlur-J fine-tuned model). Analyzing hard negative and positive examples in Figures~\ref{fig:uhdmbad} (see Appendix~\ref{appendix})  and~\ref{fig:uhdmgood} (see Appendix~\ref{appendix}), we can observe similar trend, that the improved model achieves better results in restoring details and colors both empirically and perceptually. The challenging nature of the UHDM dataset is reflected in the relatively low PSNR scores across all models, with even the best performing model achieving only 21.359 dB. This can be attributed to UHDM's high-resolution images and complex blur patterns, which pose significant challenges for deblurring models. 

\begin{table}[ht]
\centering
\caption{Performance comparison of models on UHDM~\cite{zhang_mc-blur_2023}}
\label{tab:performance_comparison}
\resizebox{\columnwidth}{!}{%
\begin{tabular}{l|c|c|c|c|c}
\hline
\textbf{Model} & \textbf{PSNR } $\uparrow$ & \textbf{SSIM } $\uparrow$ & \textbf{MAE } $\downarrow$ & \textbf{LPIPS } $\downarrow$ & \textbf{DeltaE } $\downarrow$ \\
\hline
Model checkpoint & \underline{21.323} & 0.635 & \underline{0.066} & \textbf{0.315} & \underline{3.877} \\
Reproduced Model (ours) & 21.260 & 0.618 & 0.066 & 0.345 & \textbf{3.867} \\
Fine-tuned on RealBlur-J & 20.565 & \textbf{0.664} & 0.075 & \underline{0.317} & 4.365 \\
Fine-tuned on RealBlur-R & 21.198 & \underline{0.650} & 0.068 & 0.364 & 3.966 \\
Improved Model (ours) & \textbf{21.359} & 0.637 & \textbf{0.065} & 0.325 & 4.023 \\
\hline
\end{tabular}%
}
\end{table}

\section{Ablation Studies}
In this section, we present a detailed analysis of the modifications made to the Restormer model and their impact on performance by comparing on the baseline model trained from scratch.

\begin{figure}[ht]
    \centering
    \includegraphics[width=\linewidth]{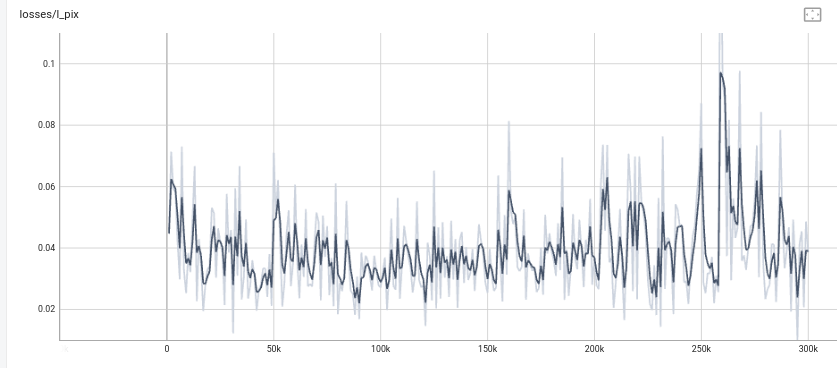}
    \caption{Tensorboard training progression of L1 loss over 300K iterations for the reproduced model.}
    \label{fig:loss_scratch}
\end{figure}

\begin{figure}[ht]
    \centering
    \includegraphics[width=\linewidth]{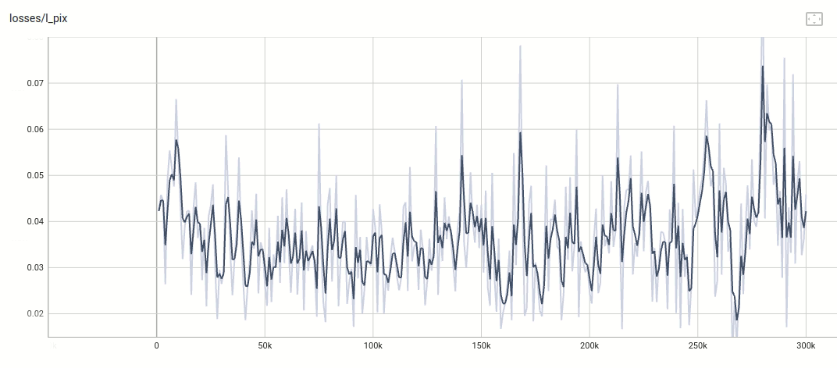}
    \caption{Tensorboard training progression of L1 loss over 300K iterations for the improved model.}
    \label{fig:loss_new}
\end{figure}

The comparison of loss, PSNR, and SSIM metrics between the two versions of the Restormer model reveals deeper insights into their training dynamics and performance. The loss curve of the baseline model (Fig.~\ref{fig:loss_scratch}) shows persistent oscillations throughout the training process, indicative of training instability, which can be attributed to the limited data augmentations and reliance solely on L1 loss. On the other hand, the loss curve of the modified model (Fig.~\ref{fig:loss_new}) is notably smoother and converges more rapidly, indicating better stability.

\begin{figure}[ht]
    \centering
    \includegraphics[width=\linewidth]{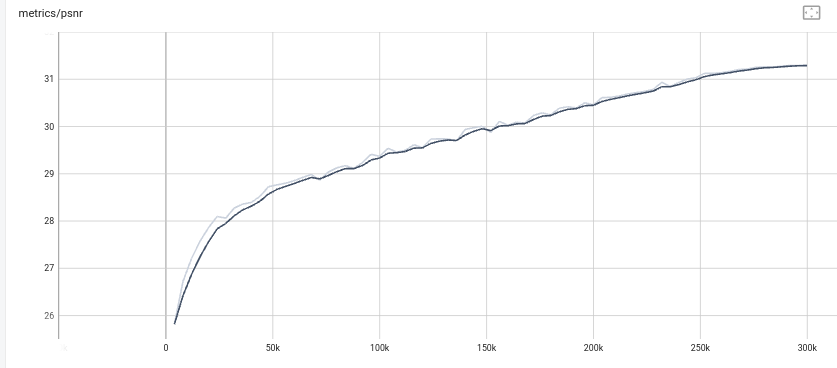}
    \caption{Tensorboard training progression of PSNR over 300K iterations for the reproduced model.}
    \label{fig:psnr_scratch}
\end{figure}

\begin{figure}[ht]
    \centering
    \includegraphics[width=\linewidth]{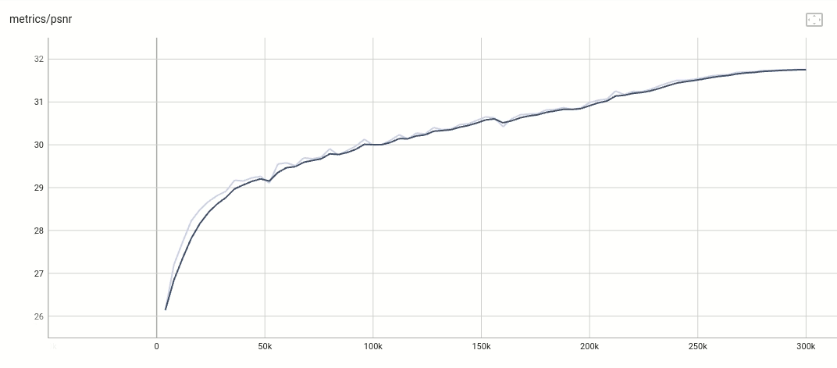}
    \caption{Tensorboard training progression of PSNR over 300K iterations for the improved model.}
    \label{fig:psnr_new}
\end{figure}

\begin{figure}[ht]
    \centering
    \includegraphics[width=\linewidth]{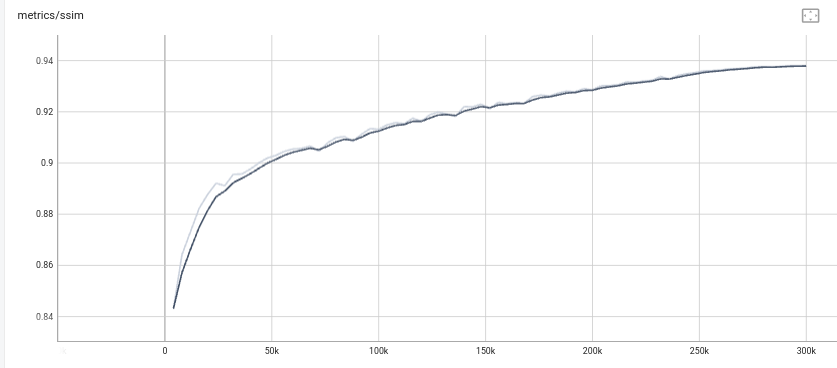}
    \caption{Tensorboard training progression of SSIM over 300K iterations for the reproduced model.}
    \label{fig:ssim_scratch}
\end{figure}

\begin{figure}[ht]
    \centering
    \includegraphics[width=\linewidth]{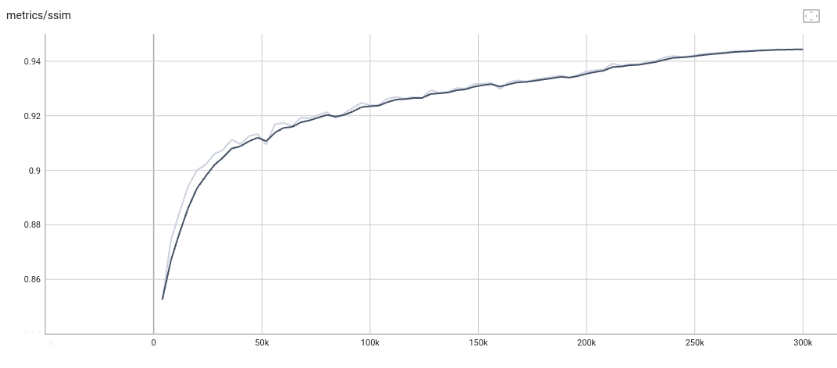}
    \caption{Tensorboard training progression of SSIM over 300K iterations for the improved model.}
    \label{fig:ssim_new}
\end{figure}

When examining the PSNR progression, the baseline model (Fig.~\ref{fig:psnr_scratch}) follows a relatively linear improvement path, plateauing around 31 dB. In comparison, the modified model's PSNR curve (Fig.~\ref{fig:psnr_new}) shows a steeper initial rise and higher final value of over 31.5 dB, reflecting faster learning and improved reconstruction quality. Similarly, the SSIM metric, which measures structural and perceptual fidelity, shows a steady increase for the baseline model (Fig.~\ref{fig:ssim_scratch}), reaching approximately 0.94, but the modified model's SSIM curve (Fig.~\ref{fig:ssim_new}) achieves higher final performance (0.945).

To better examine the impact of the modification made to the model, we test all the models on a new benchmark dataset from~\cite{zhang_mc-blur_2023} called UHDM. A good modification should lead to better results in evaluation metrics. According to the Table~\ref{tab:performance_comparison}, the improved model achieves the higher PSNR and SSIM compared to the baseline model. However, new model doesn't show better color fidelity in the reconstruction process, compared to the reproduced model which has the best result. This suggests that color transformation added to the data augmentation process might be redundant, although when benchmarking on RealBlur~\cite{vedaldi_real-world_2020} dataset, improved model shows greater performance.

The architectural changes shown in Figure~\ref{fig:arch} and ~\ref{fig:changes} are another crucial factor contributing to the improved training dynamics and performance. The reduction in parameters (18.4\%) and total layers (30\%) (see Fig.~\ref{fig:changes})  made the model more efficient while maintaining its representational power, according to Table~\ref{tab:inference_comparison}. Overall training of the baseline model took \textbf{28 hours}, while the model with reduced complexity took \textbf{23 hours}, giving a difference of 5 hours on NVIDIA RTX4090 GPU.  At the same time, the doubling of attention heads per stage enhanced the model's ability to focus on multiple regions of the image simultaneously, which likely contributed to the observed performance gains in both PSNR and SSIM. Moreover, as it was discussed before due to \textbf{GPU memory constraints}, we retrained the baseline model using \textbf{progressive training on 1 GPU}, starting with a patch size of $128 \times 128$ and a batch size of 8, and following:
\[\{(160,4), (192,4), (256,2), (320,1), (320,1)\},\]
at iterations $\{92K, 156K, 204K, 240K, 276K\}$.
In contrast, the new setup was more memory efficient allowing us to use a bigger batch size for progressive training starting with a patch size of $128 \times 128$ and a batch size of 8, and
\[\{ (160,6), (192,4), (256,2), (320,2), (384,1)\}.\] In addition to this, the changes led to the significant reduction in model size, with our model's weights totaling \textbf{81.5 MB} compared to \textbf{99.9 MB} for the original models and 99.8 MB for the fine-tuned versions highlighting the efficiency of our architecture. Overall, these adjustments balance computational efficiency and feature representation.

\begin{table}[ht]
\centering
\caption{Inference times comparison of models on the datasets.}
\label{tab:inference_comparison}
\resizebox{\columnwidth}{!}{%
\begin{tabular}{l|c|c|c}
\hline
\textbf{Model}              & \textbf{RealBlur-R (s)} & \textbf{RealBlur-J (s)} & \textbf{UHDM (s)} \\ \hline
Model checkpoint            & 625.26       & 636.69 &  3841.92 \\
Reproduced Model (ours)       & 630.87                & 638.09                 & 3698.42 \\ 
Fine-tuned                  & 631.96     & 638.33                 & 3551.56 \\ 
Improved Model (ours)       & \textbf{615.56}       & \textbf{635.51}        & \textbf{3517.01} \\ 
\hline
\end{tabular}%
}
\end{table}

Overall, the modified version of Restormer demonstrates clear advantages over the baseline, including faster convergence, smoother training curves, and superior final performance in terms of PSNR and SSIM. The improvements can be attributed to both the enhanced data augmentations, which diversified the training data, and the architectural refinements, which optimized computational efficiency and attention mechanisms. These findings underscore the effectiveness of the modifications and highlight the importance of thoughtful augmentations and architectural adjustments in achieving state-of-the-art performance.

\section{Discussion}
This work presents several important findings and insights for the field of image deblurring using transformer-based architectures. First, our results demonstrate that careful architectural optimization can significantly reduce model complexity without sacrificing performance. The 18.4\% reduction in parameters, achieved primarily through strategic reduction in transformer blocks and layers while doubling attention heads, suggests that many existing architectures may be over parameterized for their target tasks.

The improved training dynamics observed with our modified architecture are particularly noteworthy. The smoother loss curves and faster convergence indicate that our modifications not only reduce computational overhead but also create a more stable learning environment. This is likely due to the balanced trade-off between reduced layer count and increased attention capacity per layer, allowing the model to effectively capture both local and global image features.

Our experiments with enhanced data augmentation strategies reveal the importance of comprehensive training data transformation. The introduction of color jitter, Gaussian blur, and perspective transforms led to improved model robustness, particularly evident in the handling of challenging cases in real-world datasets. This suggests that careful consideration of training data variants can be as important as architectural choices for model performance.

The fine-tuning results on specific datasets demonstrate the model's adaptability, with significant improvements in both quantitative metrics and visual quality. However, the decreased performance of dataset-specific fine-tuned models on UHDM highlights an important limitation: the potential for overfitting when optimizing for specific data distributions. This trade-off between specialized and general performance remains an important consideration for practical applications.

One of the most significant contributions of this work is the demonstration that efficient transformer architectures can be developed without compromising on quality. The reduced training time and memory requirements make our approach more accessible for real-world applications, particularly in resource-constrained environments.

Future work could explore several promising directions: investigation of dynamic attention mechanisms that adapt to image content, development of more sophisticated data augmentation strategies, and exploration of hybrid architectures that combine the efficiency of our approach with other complementary techniques. Additionally, the extension of our optimization strategies to other image deblurring tasks could yield valuable insights for the broader field.

In conclusion, our work provides a practical framework for developing more efficient transformer-based image deblurring models while maintaining high performance. The principles demonstrated here have potential applications beyond motion deblurring, contributing to the broader goal of making deep learning-based image deblurring more practical and accessible.

\bibliographystyle{plain} 
\bibliography{references} 

\begin{thebibliography}{10}

\bibitem{Benjdira_Ali_Koubaa_2023}
Bilel Benjdira, Anas~M. Ali, and Anis Koubaa.
\newblock Guided frequency loss for image restoration.
\newblock (arXiv:2309.15563), October 2023.
\newblock arXiv:2309.15563 [cs, eess].

\bibitem{biyouki_comprehensive_2023}
Sajjad~Amrollahi Biyouki and Hoon Hwangbo.
\newblock A {Comprehensive} {Survey} on {Deep} {Neural} {Image} {Deblurring}, October 2023.
\newblock arXiv:2310.04719.

\bibitem{chu_improving_2022}
Xiaojie Chu, Liangyu Chen, Chengpeng Chen, and Xin Lu.
\newblock Improving {Image} {Restoration} by {Revisiting} {Global} {Information} {Aggregation}, August 2022.
\newblock arXiv:2112.04491 [cs].

\bibitem{dosovitskiy_image_2021}
Alexey Dosovitskiy, Lucas Beyer, Alexander Kolesnikov, Dirk Weissenborn, Xiaohua Zhai, Thomas Unterthiner, Mostafa Dehghani, Matthias Minderer, Georg Heigold, Sylvain Gelly, Jakob Uszkoreit, and Neil Houlsby.
\newblock An {Image} is {Worth} 16x16 {Words}: {Transformers} for {Image} {Recognition} at {Scale}, June 2021.
\newblock arXiv:2010.11929.

\bibitem{hendrycks_gaussian_2023}
Dan Hendrycks and Kevin Gimpel.
\newblock Gaussian {Error} {Linear} {Units} ({GELUs}), June 2016.
\newblock arXiv:1606.08415.

\bibitem{lee_real-time_2024}
Donghyun Lee, Hyeoksu Kwon, and Kyoungsu Oh.
\newblock Real-{Time} {Motion} {Blur} {Using} {Multi}-{Layer} {Motion} {Vectors}.
\newblock {\em Applied Sciences}, 14(11):4626, January 2024.
\newblock Number: 11 Publisher: Multidisciplinary Digital Publishing Institute.

\bibitem{loshchilov_decoupled_2019}
Ilya Loshchilov and Frank Hutter.
\newblock Decoupled {Weight} {Decay} {Regularization}, November 2017.
\newblock arXiv:1711.05101.

\bibitem{loshchilov_sgdr_2017}
Ilya Loshchilov and Frank Hutter.
\newblock {SGDR}: {Stochastic} {Gradient} {Descent} with {Warm} {Restarts}, May 2017.
\newblock arXiv:1608.03983 [cs].

\bibitem{Nah_Kim_Lee_2017}
Seungjun Nah, Tae~Hyun Kim, and Kyoung~Mu Lee.
\newblock Deep multi-scale convolutional neural network for dynamic scene deblurring.
\newblock In {\em 2017 IEEE Conference on Computer Vision and Pattern Recognition (CVPR)}, page 257–265, Honolulu, HI, July 2017. IEEE.

\bibitem{vedaldi_real-world_2020}
Jaesung Rim, Haeyun Lee, Jucheol Won, and Sunghyun Cho.
\newblock Real-{World} {Blur} {Dataset} for {Learning} and {Benchmarking} {Deblurring} {Algorithms}.
\newblock In Andrea Vedaldi, Horst Bischof, Thomas Brox, and Jan-Michael Frahm, editors, {\em Computer {Vision} – {ECCV} 2020}, volume 12370, pages 184--201. Springer International Publishing, Cham, 2020.
\newblock Series Title: Lecture Notes in Computer Science.

\bibitem{ronneberger_u-net_2015}
Olaf Ronneberger, Philipp Fischer, and Thomas Brox.
\newblock U-{Net}: {Convolutional} {Networks} for {Biomedical} {Image} {Segmentation}, May 2015.
\newblock arXiv:1505.04597.

\bibitem{psnr}
Xiwu Shang, Jie Liang, Guozhong Wang, Haiwu Zhao, Chengjia Wu, and Chang Lin.
\newblock Color-{Sensitivity}-{Based} {Combined} {PSNR} for {Objective} {Video} {Quality} {Assessment}.
\newblock {\em IEEE Transactions on Circuits and Systems for Video Technology}, 29(5):1239--1250, May 2019.
\newblock Conference Name: IEEE Transactions on Circuits and Systems for Video Technology.

\bibitem{Sharma_Wu_Dalal_2005}
Gaurav Sharma, Wencheng Wu, and Edul~N. Dalal.
\newblock The ciede2000 color-difference formula: Implementation notes, supplementary test data, and mathematical observations.
\newblock {\em Color Research \& Application}, 30(1):21–30, February 2005.

\bibitem{shi_real-time_2016}
Wenzhe Shi, Jose Caballero, Ferenc Huszár, Johannes Totz, Andrew~P. Aitken, Rob Bishop, Daniel Rueckert, and Zehan Wang.
\newblock Real-{Time} {Single} {Image} and {Video} {Super}-{Resolution} {Using} an {Efficient} {Sub}-{Pixel} {Convolutional} {Neural} {Network}.
\newblock In {\em 2016 {IEEE} {Conference} on {Computer} {Vision} and {Pattern} {Recognition} ({CVPR})}, pages 1874--1883, June 2016.
\newblock ISSN: 1063-6919.

\bibitem{Tsai_Peng_Lin_Tsai_Lin_2022}
Fu-Jen Tsai, Yan-Tsung Peng, Yen-Yu Lin, Chung-Chi Tsai, and Chia-Wen Lin.
\newblock Stripformer: Strip transformer for fast image deblurring.
\newblock In Shai Avidan, Gabriel Brostow, Moustapha Cissé, Giovanni~Maria Farinella, and Tal Hassner, editors, {\em Computer Vision – ECCV 2022}, page 146–162, Cham, 2022. Springer Nature Switzerland.

\bibitem{vaswani_attention_2023}
Ashish Vaswani, Noam Shazeer, Niki Parmar, Jakob Uszkoreit, Llion Jones, Aidan~N. Gomez, Lukasz Kaiser, and Illia Polosukhin.
\newblock Attention {Is} {All} {You} {Need}, June 2017.
\newblock arXiv:1706.03762.

\bibitem{wang_uformer_2022}
Zhendong Wang, Xiaodong Cun, Jianmin Bao, Wengang Zhou, Jianzhuang Liu, and Houqiang Li.
\newblock Uformer: {A} {General} {U}-{Shaped} {Transformer} for {Image} {Restoration}.
\newblock In {\em 2022 {IEEE}/{CVF} {Conference} on {Computer} {Vision} and {Pattern} {Recognition} ({CVPR})}, pages 17662--17672, New Orleans, LA, USA, June 2022. IEEE.

\bibitem{ssim}
Zhou Wang, A.C. Bovik, H.R. Sheikh, and E.P. Simoncelli.
\newblock Image quality assessment: from error visibility to structural similarity.
\newblock {\em IEEE Transactions on Image Processing}, 13(4):600--612, April 2004.
\newblock Conference Name: IEEE Transactions on Image Processing.

\bibitem{Yoshihara_Fukiage_Nishida_2023}
Sou Yoshihara, Taiki Fukiage, and Shin’ya Nishida.
\newblock Does training with blurred images bring convolutional neural networks closer to humans with respect to robust object recognition and internal representations?
\newblock {\em Frontiers in Psychology}, 14:1047694, February 2023.

\bibitem{zamir_restormer_2022}
Syed~Waqas Zamir, Aditya Arora, Salman Khan, Munawar Hayat, Fahad~Shahbaz Khan, and Ming-Hsuan Yang.
\newblock Restormer: {Efficient} {Transformer} for {High}-{Resolution} {Image} {Restoration}, March 2022.
\newblock arXiv:2111.09881.

\bibitem{zamir_multi-stage_2021}
Syed~Waqas Zamir, Aditya Arora, Salman Khan, Munawar Hayat, Fahad~Shahbaz Khan, Ming-Hsuan Yang, and Ling Shao.
\newblock Multi-{Stage} {Progressive} {Image} {Restoration}, March 2021.
\newblock arXiv:2102.02808.

\bibitem{zhang_deep_2022}
Kaihao Zhang, Wenqi Ren, Wenhan Luo, Wei-Sheng Lai, Bjorn Stenger, Ming-Hsuan Yang, and Hongdong Li.
\newblock Deep {Image} {Deblurring}: {A} {Survey}, May 2022.
\newblock arXiv:2201.10700.

\bibitem{zhang_mc-blur_2023}
Kaihao Zhang, Tao Wang, Wenhan Luo, Boheng Chen, Wenqi Ren, Bjorn Stenger, Wei Liu, Hongdong Li, and Ming-Hsuan Yang.
\newblock {MC}-{Blur}: {A} {Comprehensive} {Benchmark} for {Image} {Deblurring}, September 2023.
\newblock arXiv:2112.00234.

\bibitem{lpips}
Richard Zhang, Phillip Isola, Alexei~A. Efros, Eli Shechtman, and Oliver Wang.
\newblock The {Unreasonable} {Effectiveness} of {Deep} {Features} as a {Perceptual} {Metric}, April 2018.
\newblock arXiv:1801.03924.

\bibitem{Zhao_Wei_He_Lu_2022}
Wenda Zhao, Fei Wei, You He, and Huchuan Lu.
\newblock United defocus blur detection and deblurring via adversarial promoting learning.
\newblock In Shai Avidan, Gabriel Brostow, Moustapha Cissé, Giovanni~Maria Farinella, and Tal Hassner, editors, {\em Computer Vision – ECCV 2022}, page 569–586, Cham, 2022. Springer Nature Switzerland.

\bibitem{zheng_uhd_2022}
Zhuoran Zheng and Xiuyi Jia.
\newblock {UHD} {Image} {Deblurring} via {Multi}-scale {Cubic}-{Mixer}, June 2022.
\newblock arXiv:2206.03678 [cs].

\end{thebibliography}

\onecolumn

\appendix
\label{appendix}

This appendix presents remaining results and examples from three different datasets: RealBlur-R, RealBlur-J, and UHDM are provided. For each dataset, we also show both hard negative and hard positive examples to illustrate the challenging cases encountered during evaluation.

\section{Dataset Examples}

\begin{figure}[ht]
    \centering
    \includegraphics[width=0.7\linewidth]{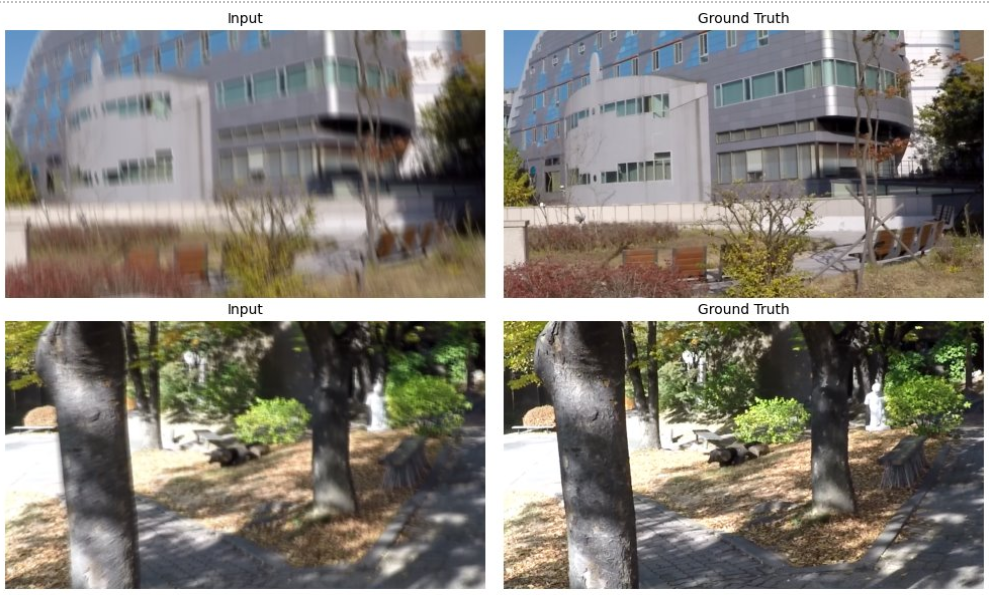}
    \caption{GoPro~\cite{Nah_Kim_Lee_2017}}
    \label{fig:gopro}
\end{figure}

\subsection{GoPro Dataset Examples in Figure~\ref{fig:gopro}.}
\begin{figure}[ht]
    \centering
    \includegraphics[width=0.6\linewidth]{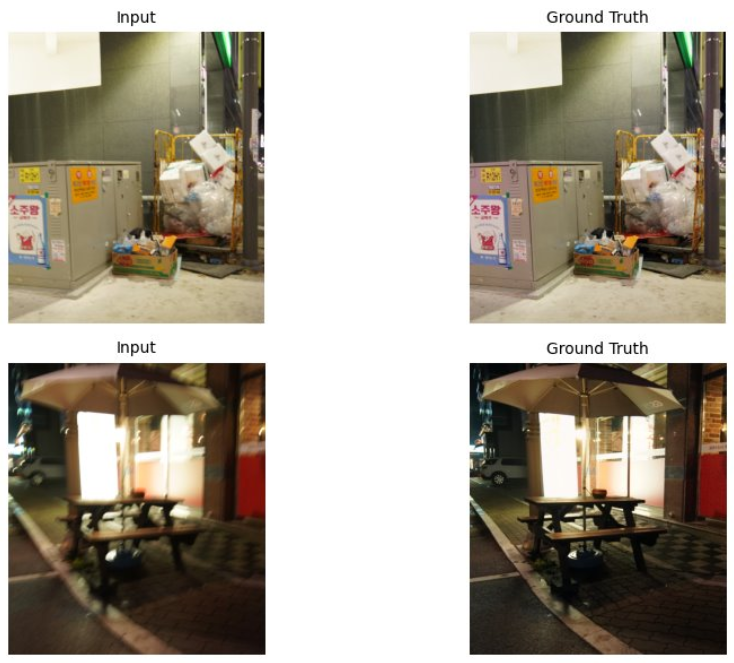}
    \caption{RealBlur~\cite{vedaldi_real-world_2020}}
    \label{fig:realblur}
\end{figure}

\subsection{ReaBlur Dataset Examples in Figure~\ref{fig:realblur}.}
\begin{figure}[ht]
    \centering
    \includegraphics[width=0.7\linewidth]{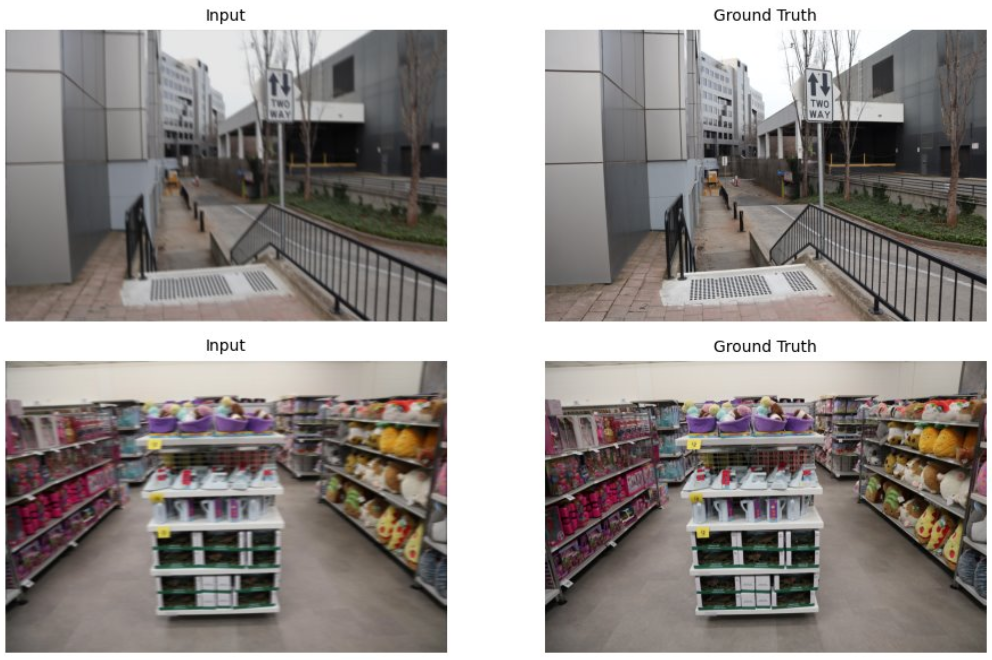}
    \caption{UHDM~\cite{zhang_mc-blur_2023}}
    \label{fig:uhdm}
\end{figure}

\subsection{UHDM Dataset Examples in Figure~\ref{fig:uhdm}.}
\subsection{RealBlur-R Dataset Examples}
\begin{figure*}[!t]
\centering
\includegraphics[height = 5cm, width=0.9\linewidth]{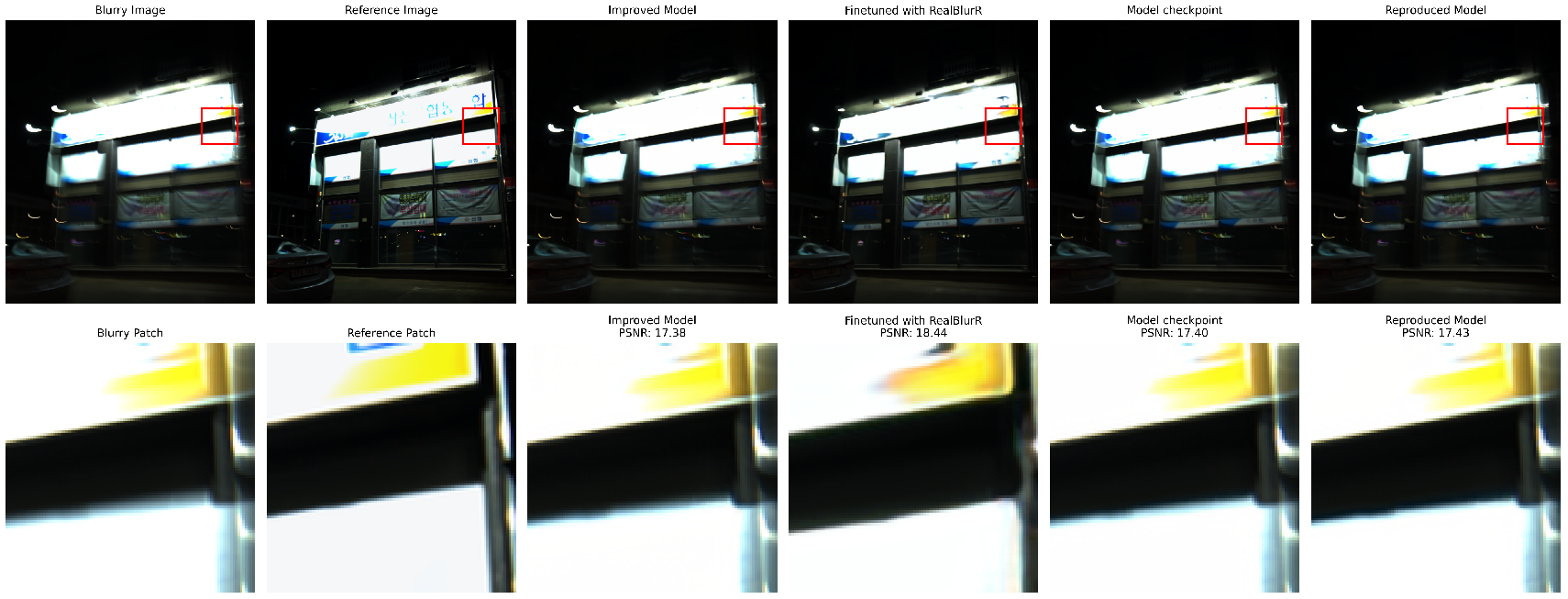}
\caption{Hard Negative example from RealBlur-J dataset (zoom in for better visibility)}
\label{fig:blurRbad}
\end{figure*}

\begin{figure*}[!t]
\centering
\includegraphics[height = 5cm, width=0.9\linewidth]{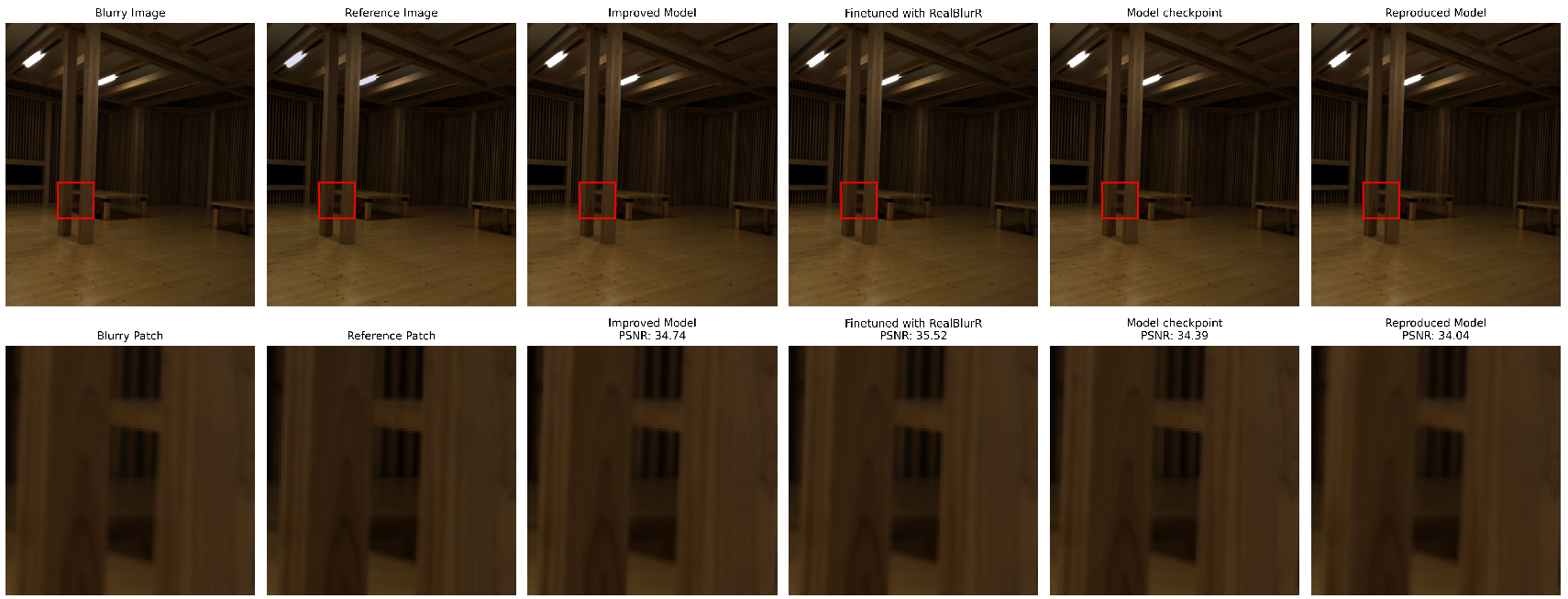}
\caption{Hard Positive example from RealBlur-J dataset (zoom in for better visibility)}
\label{fig:blurRgood}
\end{figure*}

Fig.\ref{fig:blurRbad} shows a hard negative example from the RealBlur-R dataset, while Fig.\ref{fig:blurRgood} presents a hard positive example from the same dataset. These examples demonstrate the subtle differences that make classification challenging in real-world scenarios.
\subsection{RealBlur-J Dataset Examples}
\begin{figure*}[!t]
\centering
\includegraphics[height = 5cm, width=0.9\linewidth]{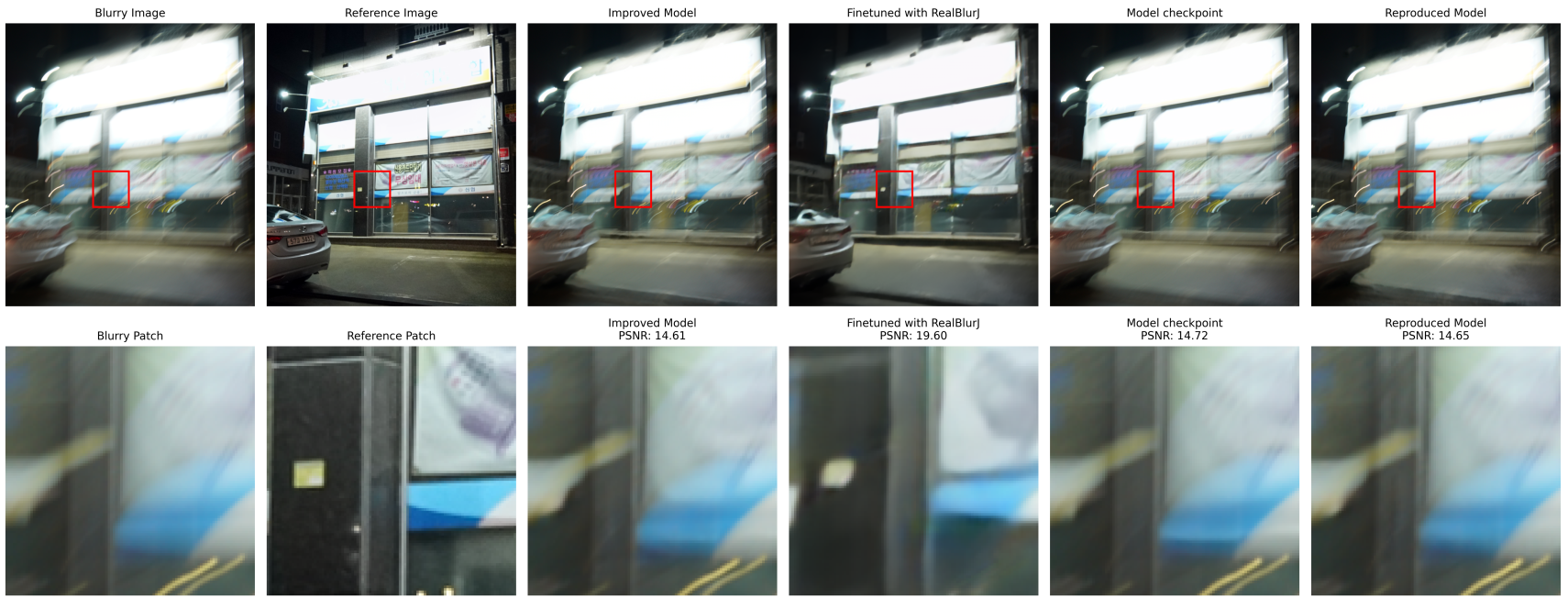}
\caption{Hard Negative example from RealBlur-J~\cite{vedaldi_real-world_2020} dataset (zoom in for better visibility)}
\label{fig:blurJbad}
\end{figure*}

\begin{figure*}[!t]
\centering
\includegraphics[height = 5cm, width=0.9\linewidth]{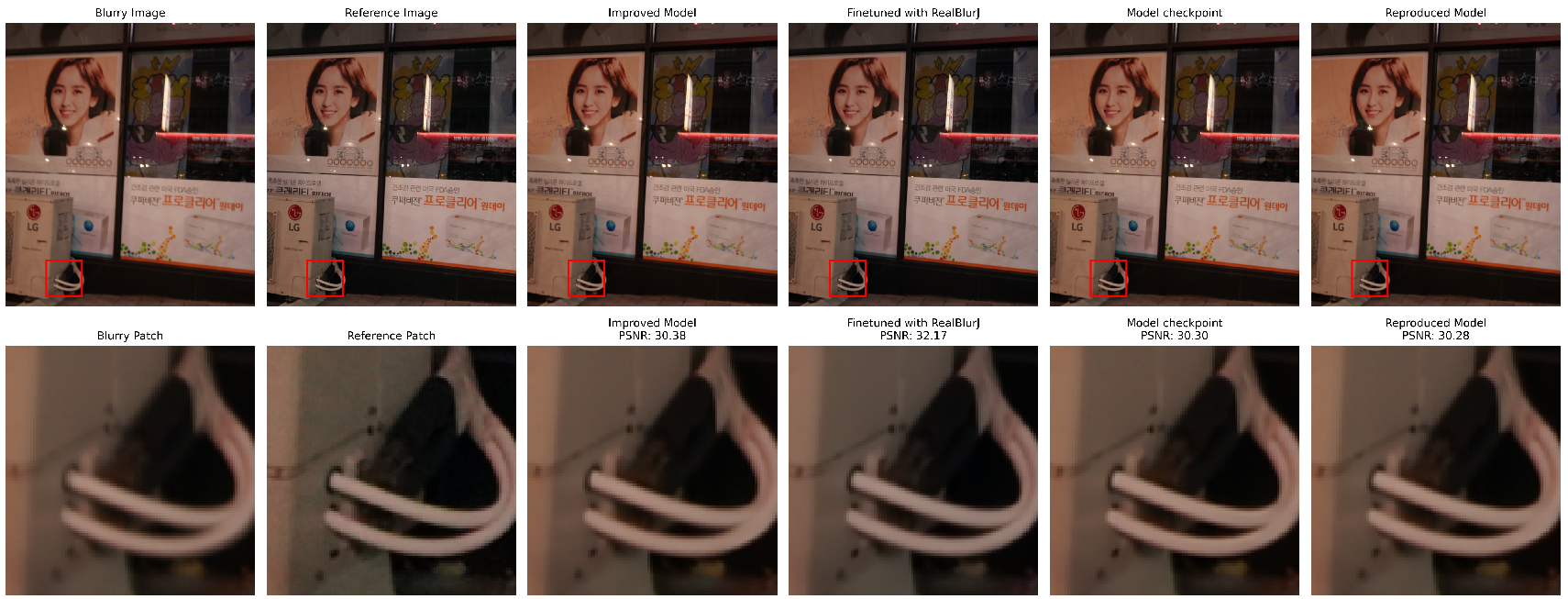}
\caption{Hard Positive example from RealBlur-J~\cite{vedaldi_real-world_2020} dataset (zoom in for better visibility)}
\label{fig:blurJgood}
\end{figure*}

Fig.\ref{fig:blurJbad} and Fig.\ref{fig:blurJgood} showcase hard negative and positive examples respectively from the RealBlur-J dataset~\cite{vedaldi_real-world_2020}. These examples highlight the complexity of blur detection in diverse imaging conditions.
\subsection{UHDM Dataset Examples}
\begin{figure*}[!t]
\centering
\includegraphics[height = 5cm,width=1\linewidth]{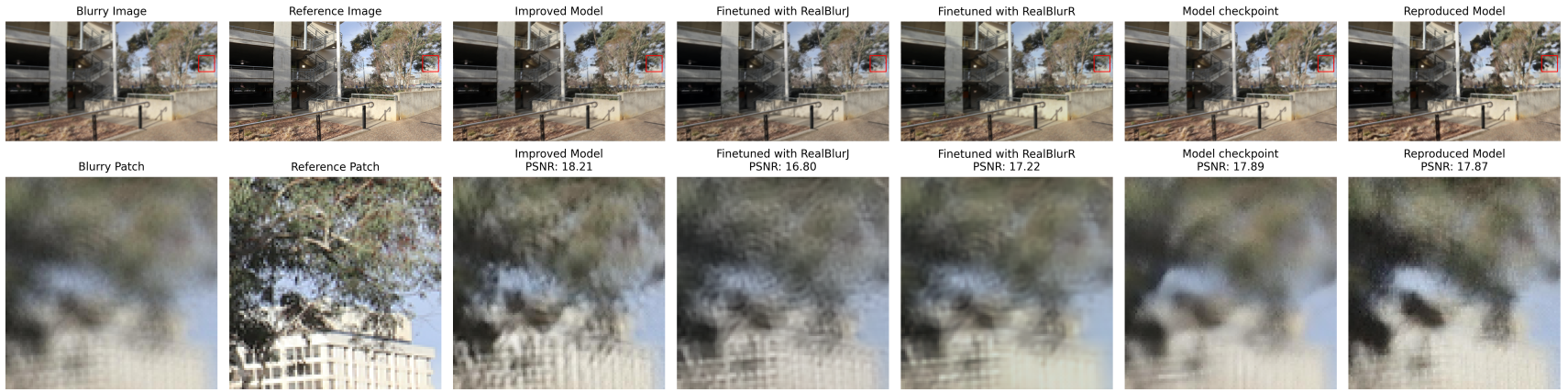}
\caption{Hard Negative example from UHDM~\cite{zhang_mc-blur_2023} dataset (zoom in for better visibility)}
\label{fig:uhdmbad}
\end{figure*}

\begin{figure*}[!t]
\centering
\includegraphics[height = 5cm, width=1\linewidth]{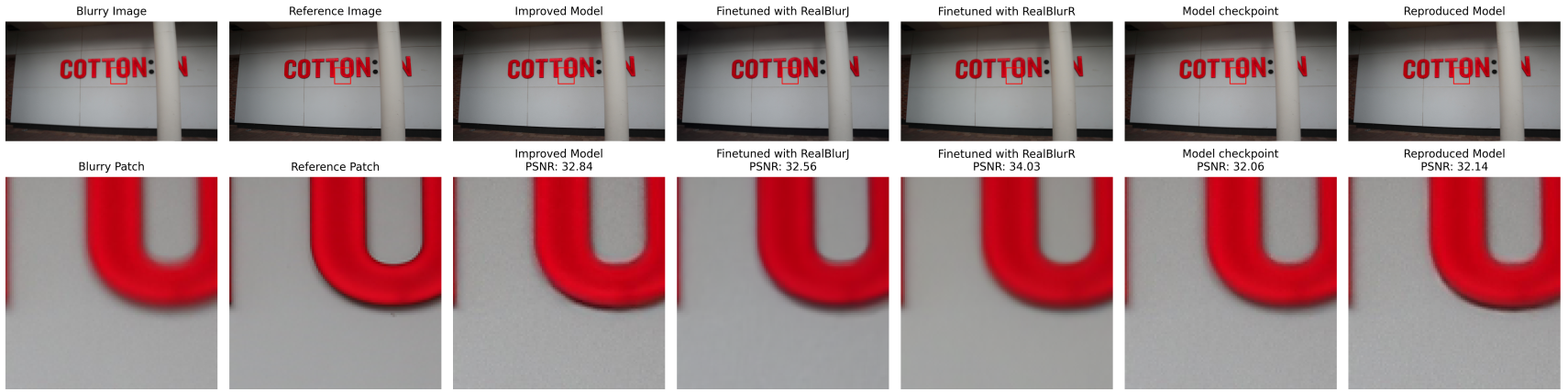}
\caption{Hard Positive example from UHDM~\cite{zhang_mc-blur_2023} dataset (zoom in for better visibility)}
\label{fig:uhdmgood}
\end{figure*}

The UHDM dataset~\cite{zhang_mc-blur_2023} examples are presented in Fig.\ref{fig:uhdmbad} and Fig.\ref{fig:uhdmgood}, showing hard negative and positive cases respectively. These examples demonstrate the challenges in blur detection across different resolutions and scene types.

\onecolumn

\end{document}